\documentclass[11pt,a4paper]{article}
\usepackage[hyperref]{emnlp2018}
\usepackage{times}
\usepackage{latexsym}
\usepackage{graphicx}
\usepackage{algorithm,algorithmicx,algpseudocode}
\usepackage{hyperref,etoolbox}
\usepackage{amsmath}
\usepackage{amsfonts}
\usepackage{amssymb}
\usepackage{latexsym}
\usepackage{multirow}
\usepackage{xspace}
\usepackage{url}
\usepackage{color}

\usepackage{stackengine}
\usepackage{enumitem}
\usepackage{booktabs}
\usepackage{soul}

\stackMath

\newcommand{\defeq}{\triangleq}

\newcommand{\nop}[1]{}

\newcommand{\expect}[1]{\ensuremath{\underset{#1}{\mathbb{E}}\xspace}}
\allowdisplaybreaks

\aclfinalcopy 

\title{XL-NBT: A Cross-lingual Neural Belief Tracking Framework}

\author{Wenhu Chen$^{1}$, Jianshu Chen$^{2}$, Yu Su$^{3}$, Xin Wang$^{1}$, Dong Yu$^{2}$, Xifeng Yan$^{1}$ and William Wang$^{1}$\\
  University of California, Santa Barbara, CA, USA$^1$\\ 
  Tencent AI Lab, Bellevue, WA, USA$^2$\\
  The Ohio State University, Columbus, Ohio$^3$\\
 \texttt{\{wenhuchen, xwang, xyan, william\}@cs.ucsb.edu, su.809@osu.edu}\\
 \texttt{jianshuchen@tencent.com, dongyu@ieee.org}\\
}

\date{}

\begin{document}
\maketitle
\begin{abstract}
Task-oriented dialog systems are becoming pervasive, and many companies heavily rely on them to complement human agents for customer service in call centers. With globalization, the need for providing cross-lingual customer support becomes more urgent than ever. However, cross-lingual support poses great challenges---it requires a large amount of additional annotated data from native speakers. In order to bypass the expensive human annotation and achieve the first step towards the ultimate goal of building a universal dialog system, we set out to build a cross-lingual state tracking framework. Specifically, we assume that there exists a source language with dialog belief tracking annotations while the target languages have no annotated dialog data of any form. Then, we pre-train a state tracker for the source language as a teacher, which is able to exploit easy-to-access parallel data. We then distill and transfer its own knowledge to the student state tracker in target languages. We specifically discuss two types of common parallel resources: bilingual corpus and bilingual dictionary, and design different transfer learning strategies accordingly. Experimentally, we successfully use English state tracker as the teacher to transfer its knowledge to both Italian and German trackers and achieve promising results.
\end{abstract}

\section{Introduction}
Over the past few years, we have witnessed the burgeoning of real-world applications of dialog systems, with many academic, industrial, and startup efforts racing to lead the widely-believed next-generation human-machine interfaces.
As a result, numerous task-oriented dialog systems such as virtual assistants and customer conversation services were  developed~\cite{DBLP:conf/emnlp/WenGMSVY15,DBLP:conf/eacl/Rojas-BarahonaG17,BordesW16,DBLP:conf/acl/WilliamsAZ17,DBLP:conf/ijcnlp/LiCLGC17}, with Google Duplex\footnote{\url{https://ai.googleblog.com/2018/05/duplex-ai-system-for-natural-conversation.html}} being the most recent example.

With the rapid process of globalization, more countries have observed growing populations of immigrants, 
and more companies have moved forward to develop their overseas business sectors. To provide better customer service and bring down the cost of labor at call centers, the development of universal dialog systems has become a practical issue. A straightforward strategy is to separately collect training data and train dialog systems for each language. However, it is not only tedious but also expensive. Two settings naturally arise for more efficient usage of the training data: (1) Multi-lingual setting: we annotate data for multiple languages and train a single model, with possible innovations on joint training. (2) Cross-lingual setting: we annotate data and train a model for only one (popular) language, and transfer the learned knowledge to other languages. Here we are interested in the second case, and the important research question we ask is: 
\emph{How can we build cross-lingual dialog systems that can support less popular, low- or even zero-resource languages}? 

As an initial step towards cross-lingual dialog systems, we focus on the cornerstone of dialog systems -- \emph{dialog state tracking} (DST), or \emph{belief tracking}, a key component for understanding user inputs and updating \emph{belief state}, i.e., a system's internal representation of the state of conversation \cite{young2010hidden}.
Based on the perceived belief state, the dialog manager can decide which action to take, and what verbal response to generate~\cite{DBLP:conf/icml/2017,BordesW16}. 

\begin{figure*}[!thb]
    \begin{center}
    \includegraphics[width=1.0\linewidth]{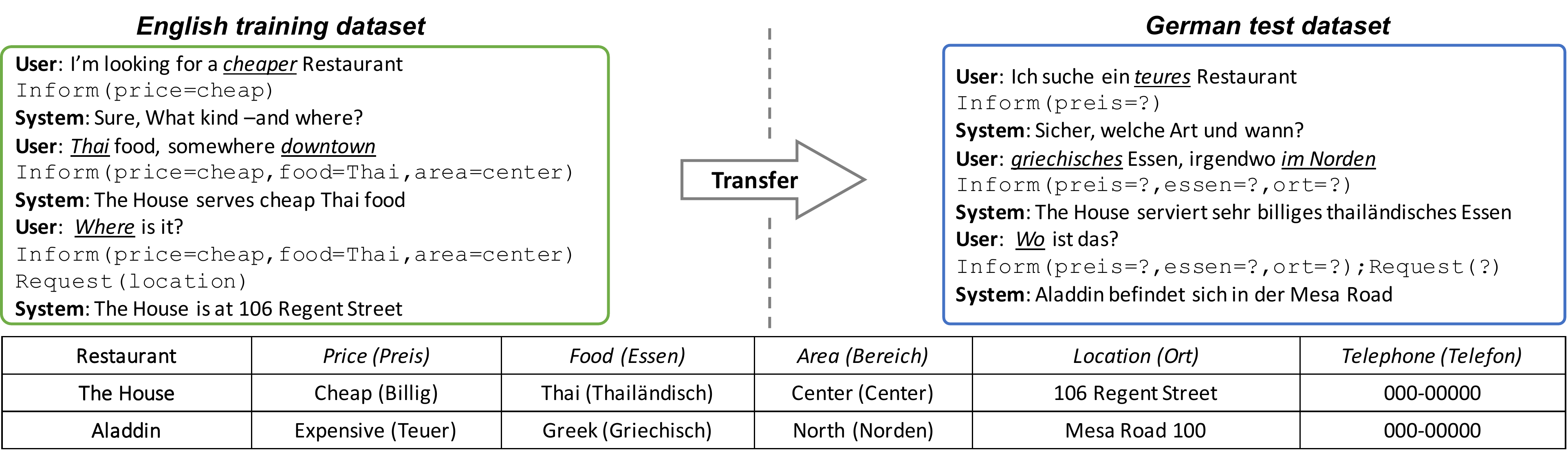}
    \end{center}
    \vspace*{-2ex}
    \caption{Cross-lingual transfer learning for dialog state tracking, where the underlying database (the table above) is shared across languages. The source language has annotated dialogs and the ground truth states, but the target language has neither dialogs nor ground truth states (only a testing dataset for evaluation).}
    \label{fig:overview}
    \vspace*{-2ex}
\end{figure*}

DST models require a considerable amount of annotated data for training \cite{DBLP:conf/sigdial/HendersonTY14,DBLP:conf/acl/MrksicSTGSVWY15,DBLP:conf/acl/MrksicSWTY17}.
For a common dialog shown in \autoref{fig:overview}, a typical data acquisition process~\cite{DBLP:conf/eacl/Rojas-BarahonaG17} not only requires two human users to converse for multiple turns but also requires annotators to identify user's intention in each turn. Such two-step annotation is very expensive, especially for rare languages.

We study the novel problem of \emph{cross-lingual DST}, where one leverages the annotated data of a source language to train DST for a target language with \emph{zero} annotated data (\autoref{fig:overview}); no conversation dialog or dialog state annotation is available for the target language. In order to deal with this zero-resource challenging scenario, we first decouple the state-of-the-art neural belief tracker framework~\cite{DBLP:conf/acl/MrksicSWTY17} into sub-modules, namely \emph{utterance encoder}, \emph{context gate}, and \emph{slot-value decoder}. By introducing a teacher-student framework, we are able to transfer knowledge across languages module by module, following the divide-and-conquer philosophy. Requiring no target-side dialog data, our method relies on other easy-to-access parallel resources to understand the connection between languages. Depending on the popularity and availability of target language resources, we study two kinds of parallel data: bilingual corpus and bilingual dictionary, and we respectively design two transfer learning strategies.

We use the popular \textit{Wizard-of-Oz}~\cite{DBLP:conf/eacl/Rojas-BarahonaG17} dataset as our DST benchmark to evaluate the effectiveness of our cross-lingual transfer learning. We specify English as the source (primary) language and two different European languages (German and Italian) as our zero-annotation target languages. Compared with an array of alternative transfer learning strategies, our cross-lingual DST models consistently achieve promising results in both scenarios for both zero-annotation languages. To ensure reproducibility, we release our code, training data and parallel resources in the github\footnote{\small\url{https://github.com/wenhuchen/Cross-Lingual-NBT}}. Our main contributions are three-fold:
\begin{itemize}[noitemsep,topsep=0pt]
\item Towards building cross-lingual dialog systems, we are the first to study the cross-lingual dialog state tracking problem.
\item We systematically study different scenarios for this problem based on the availability of parallel data and propose novel transfer learning methods to tackle the problem.
\item We empirically demonstrate the efficacy of the proposed methods, showing that our methods can accurately track dialog states for languages with zero annotated data. 
\end{itemize}

\section{Related Work}
\subsection{Dialog State Tracking}
Broadly speaking, the dialog belief tracking algorithms can be divided into three families: 1) hand-crafted rules 2) generative models, and 3) maximum-entropy model~\cite{metallinou2013discriminative}. Later on, many deep learning based discriminative models have surged to replace the traditional strategies~\cite{henderson2014robust,DBLP:conf/acl/MrksicSWTY17,williams2016dialog} and achieved state-of-the-art results on various datasets. Though the discriminative models are reported to achieve fairly high accuracy, their applications are heavily restricted by the domain, ontology, and language. Recently, a pointer network based algorithm~\cite{DBLP:conf/acl/XuH18} and another multi-domain algorithm~\cite{DBLP:conf/asru/RastogiHH17} have been proposed to break the ontology and domain boundary. Besides, ~\cite{mrkvsic2017semantic} has proposed an algorithm to train a unified framework to deal with multiple languages with annotated datasets. In contrast, our paper focuses on breaking the language boundary and transfer DST knowledge from one language into other zero-annotation languages.
\subsection{Cross-Lingual Transfer Learning}
Cross-lingual transfer learning has been a very popular topic during the years, which can be seen as a transductive process. In such process, the input domains of the source and target are different~\cite{pan2010survey} since each language has its own distinct lexicon. By discovering the underlying connections between the source and target domain, we could design transfer algorithms for different tasks. Recently, algorithms have been successfully designed for POS tagging~\cite{DBLP:conf/naacl/2016,kim2017cross}, NER~\cite{pan2017cross,DBLP:conf/acl/NiDF17} as well as image captioning~\cite{miyazaki2016cross}. These methods first aim at discovering the relatedness between two languages and separate language-common modules from language-specific modules, then resort to external resources to transfer the knowledge across the language boundary. Our method addresses the transfer learning using a teacher-student framework and proposes to use the teacher to gradually guide the student to make more proper decisions.

\section{Problem Definition}
\begin{figure}[htb]
\begin{center}
\includegraphics[width=1.0\linewidth]{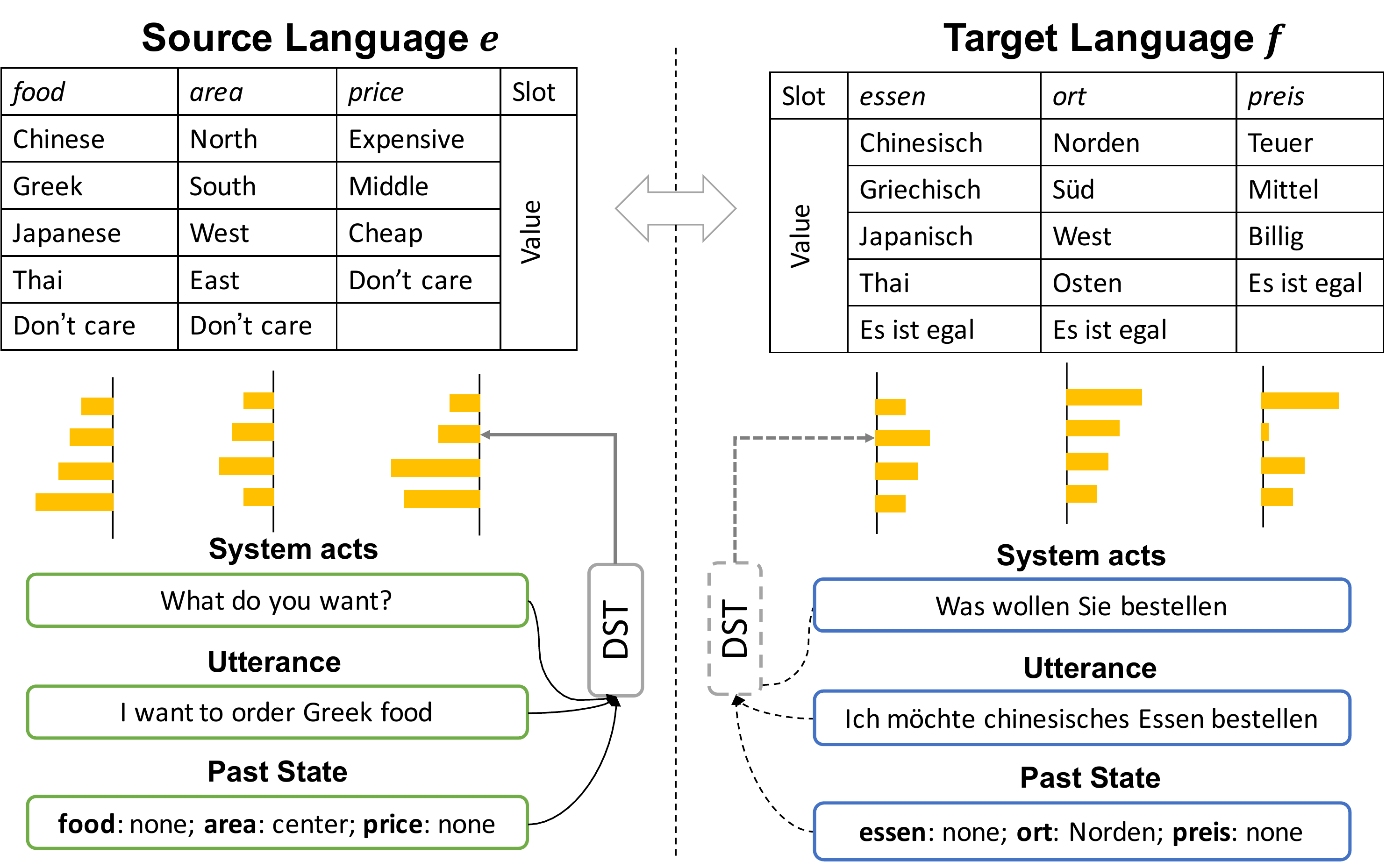}
\end{center}
    \vspace*{-2ex}
   \caption{Cross-lingual DST structure, the ontology and database between multiple languages are shared.}
\label{fig:DST}
    \vspace*{-2ex}
\end{figure}
\label{par:cross-DST}
The dialog states are defined as a set of search constraints (i.e. informable slots or goals) that the user specified through the dialog and a set of attribute questions regarding the search results (i.e. requestable slots or requests). The objective of dialog state tracking (DST) is to predict and track the user intention (i.e., the values of the aforementioned slots) at each time step based on the current user utterance and the entire dialog history. As shown in~\autoref{fig:DST}, for each slot, the DST computes an output distribution of the candidate values using three inputs: (i) \textit{system response} $a_t$, which is the sentence generated by the system, (ii) \textit{utterance} $u_{t}$, which is the sentence from the user, and (iii) \textit{previous state}, which denotes the selected slot-value pairs. We define the ontology of the dialog system to be the set of all the possible words the dialog slot and value can take. In this paper, we are interested in learning a cross-lingual DST. Specifically, we assume that the DST for the \emph{source} language has access to a human-annotated training dataset $D$ while the DSTs for the \emph{target} languages do not have access to annotated data in other languages except for testing data. We here mainly consider two different types of parallel resources to assist the transfer learning:\\
\noindent \textbf{(1) Bilingual Corpus}, where abundant bilingual corpora exist between the source and the target languages. This is often the case for common language pairs like German, Italian, and French, etc.\\
\noindent \textbf{(2) Bilingual Dictionary}, where public bilingual dictionaries exist between the source and the target languages, but large-scaled parallel corpus are harder to obtain. This can be the case for rarer languages like Finnish, Bulgarian, etc.\\
Furthermore, we assume that all the languages share a common multi-lingual database, whose column/row names and entry values are stored via multiple languages (see the database in~\autoref{fig:overview}). That is, the ontology of dialog among different languages is known with a one-to-one mapping between them (e.g., greek=griechisch=greco, food=essen=cibo). Based on that, we could construct a mapping function $M$ to associate the ontology terms from different languages with pre-designed language-agnostic concepts: for example, $M(foods)$ = $M(Essen)$ = $M(Cibo)$ = \textsc{food}. We illustrate our problem definition in~\autoref{fig:DST}.

\section{Decoupled Neural Belief Tracker}
\begin{figure}[!thb]
\begin{center}
\includegraphics[width=1.0\linewidth]{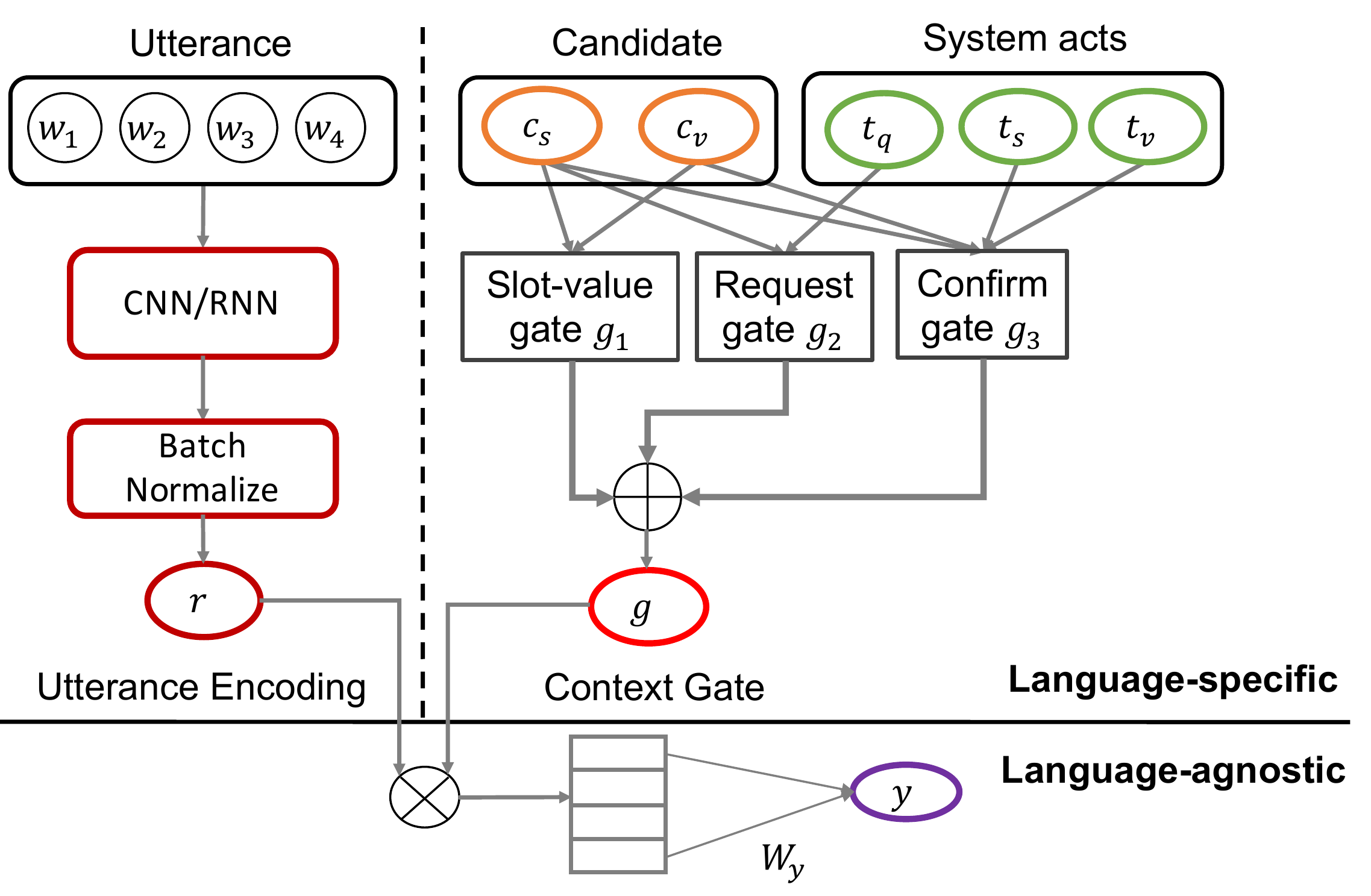}
\end{center}
    \vspace*{-2ex}
   \caption{Our implementation of baseline NBT, slightly modified from~\cite{DBLP:conf/acl/MrksicSWTY17}.}
\label{fig:NBT}
    \vspace*{0ex}
\end{figure}

We design our cross-lingual DST on top of the state-of-the-art Neural Belief Tracker (NBT)~\cite{DBLP:conf/acl/MrksicSWTY17}, which demonstrates many advantages (no hand-crafted lexicons, no linguistic knowledge required, etc). These nice properties are essential for our cross-lingual DST design because we are pursuing a general and simple framework regardless of the language properties. In short, NBT consists of a neural network that computes the matching score for every candidate slot-value pair $(c_s, c_v)$ based on the following three inputs: (i) the system dialog acts $a_t = (t_q, t_s, t_v)$,\footnote{\small $t_q$ represents the system request, $t_s, t_v$ represents the system confirmation. If the system wants to request some information from the user by asking ``what's your favorite area?'', then NBT sets $t_q$=``AREA''. If the system wants to confirm some information from a user by asking ``should I try Persian restaurants in the north?" then NBT sets $t_s, t_v$=``area, north''.} (ii) the user utterance $u_t$, and (iii) the candidate slot-value pair. And it identifies the user intents by evaluating the scores for all the slot-value pairs (see~\autoref{fig:NBT}). With a slight abuse of notation, we still use $c_s, c_v, t_s, t_v, t_q \in \mathbb{R}^H$ to denote the vector representations of themselves, where $H$ is the embedding dimension. We will use pre-trained embedding vectors in our cross-lingual NBT, just like the original NBT and they will be fixed during training. To enable cross-lingual transfer learning, we first re-interpret the architecture of the original NBT by decomposing it into three components:

\paragraph{Utterance Encoding}
The first component is an utterance encoder, which maps the utterance $u_t = \{w_1, w_2, \cdots, w_N\}$ of a particular language into a semantic representation vector $r(u_t) \in \mathbb{R}^H$, where $w_i \in \mathbb{R}^H$ is the word vector for the $i$-th token and $N$ is the length of the utterance. Note that the dimension of the semantic vector $r(u_t)$ is the same as that of the word vector.
We implement the encoder using the same convolutional neural network (CNN) as the original NBT, with a slight modification of adding a top batch normalization layer. We will explain this change in~\autoref{sec:xl-nbt}.

\paragraph{Context Gate}
The second part is the context gate, which takes the system acts $a_t=(t_q, t_s, t_v)$ and the candidate slot-value pair $(c_s,c_v)$ as its inputs and filter out the desired information from the encoded utterance. The context gate $g$ is a sum of three separate gates:
\vspace{-1ex}
\begin{align}
\begin{split}
        g(c_s, c_v, a_t) = g_1 + g_2 + g_3
\end{split}
\end{align}
where the individual gates are defined as:
\vspace{-1ex}
\begin{align}
\begin{split}
    &g_1 = \sigma(W^s_c(c_s + c_v) + b^s_c) \\
    &g_2 = (c_s \cdot W^q_t t_q) \odot [1,\cdots, 1]^H\\
    &g_3 = (c_s \cdot W^s_t t_s)(c_v \cdot W^v_t t_v) \odot [1,\cdots, 1]^H
\end{split}
\end{align}
where $W^s_c, W^q_t, W^s_t, W^v_t \in \mathbb{R}^{H \times H}$ are the weight matrices,  and $\odot$ and $\cdot$ denote the Hadamard product and the inner product, respectively. The three gates $g_1 \in \mathbb{R}^H, g_2 \in \mathbb{R}^H, g_3 \in \mathbb{R}^H$ model the relevance between the candidate slot and value, the system request and the system confirms, respectively. The transformation matrices $W^q_t, W^s_t, W^v_t$ are added to the original NBT to increase the model flexibility of the gates.

\paragraph{Slot-Value Decoding}
The final component is a slot-value decoder, which predicts the score $y$ of a given slot-value pair using the filtered information from the utterance representation $r$ as:
\vspace{-1ex}
\begin{align}
\label{eq:decompose}
\begin{split}
    &y(c_s, c_v, u_t, a_t) =  W_y^T  [r(u_t) \odot g(c_s, c_v, a_t)]
\end{split}
\end{align}
where $W_y \in \mathbb{R}^{H \times 1}$ is the weight vector. The above expression computes the score for the slot-value pair based on the information from the current turn. We combine it with the information from previous turns to get the final score:
\begin{align}
\small
\begin{split}
    \hat{y}(c_v|u_t, a_t, c_s) =& \lambda y(c_s, c_v, u_t, a_t) +\\
                      &(1 - \lambda) \hat{y}(c_s, c_v, u_{t-1}, a_{t-1})    
\end{split}
\end{align}
here $\lambda$ is a combination weight. For each given slot $c_s$, NBT selects the single highest value for informable slots and selects all values above a certain threshold for request slots. Here we replace the multi-layer perceptron in the orginal NBT by a linear output layer (to be explained in~\autoref{sec:xl-nbt}).

\section{Cross-lingual Neural Belief Tracker}
\label{sec:xl-nbt}
In this section, we develop a cross-lingual Neural Belief  Tracker (XL-NBT) that distills knowledge from one NBT to another using a teacher-student framework. We assume the ontology mapping $M$ is known a priori (see~\autoref{par:cross-DST}). XL-NBT uses language-specific utterance encoder and context gate for each input language while sharing a common (language-agnostic) slot-value decoder across different languages (see Figure \ref{fig:NBT}). The key idea is to optimize the language-specific components of the student network (NBT of the target language) so that their outputs are language-agnostic. This is achieved by making these outputs close to that of the teacher network (NBT of the source language), as we detail below.

\begin{figure*}[!thb]
\begin{center}
\includegraphics[width=1.0\linewidth]{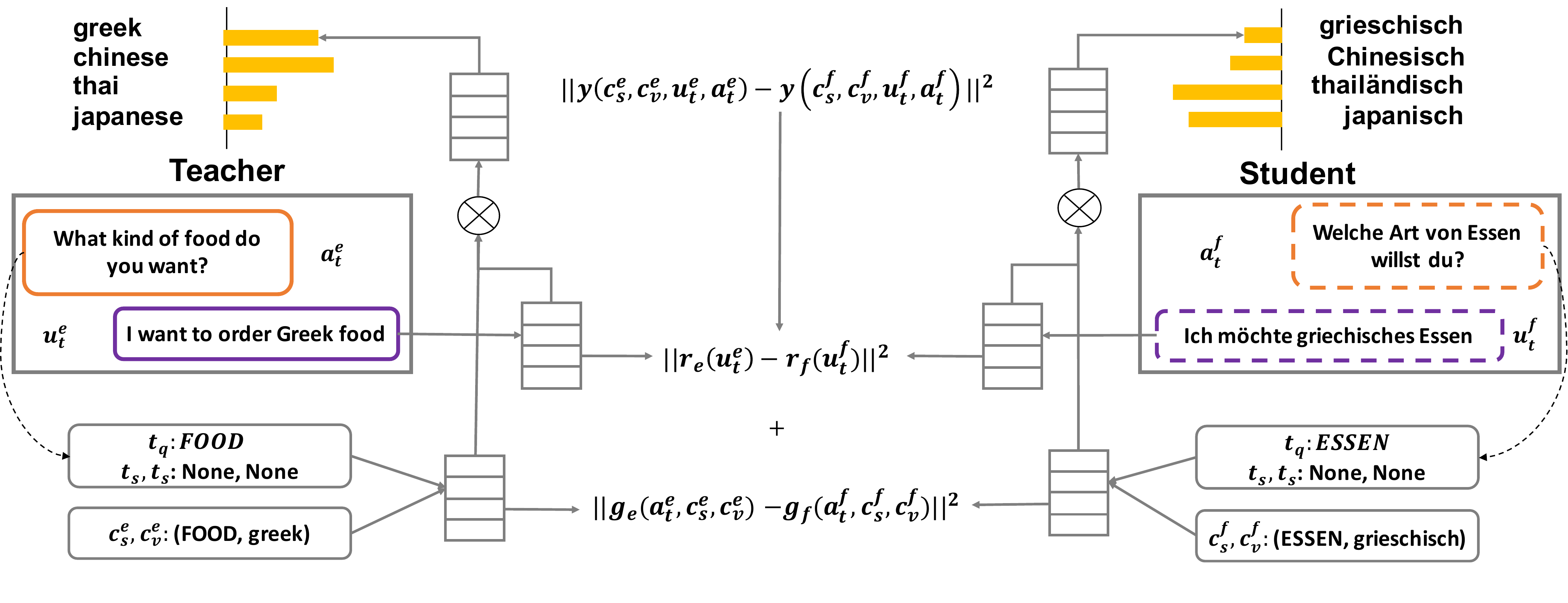}
\end{center}
    \vspace*{-2ex}
   \caption{Teacher-Student Framework for cross-lingual transfer learning. The dotted line denotes the imaginary utterances, which expresses the same intention as the source side.}
\label{fig:teacher-student}
    \vspace*{-2ex}
\end{figure*}

\subsection{Teacher-Student Framework}
\begin{figure}[thb]
\begin{center}
\includegraphics[width=1.0\linewidth]{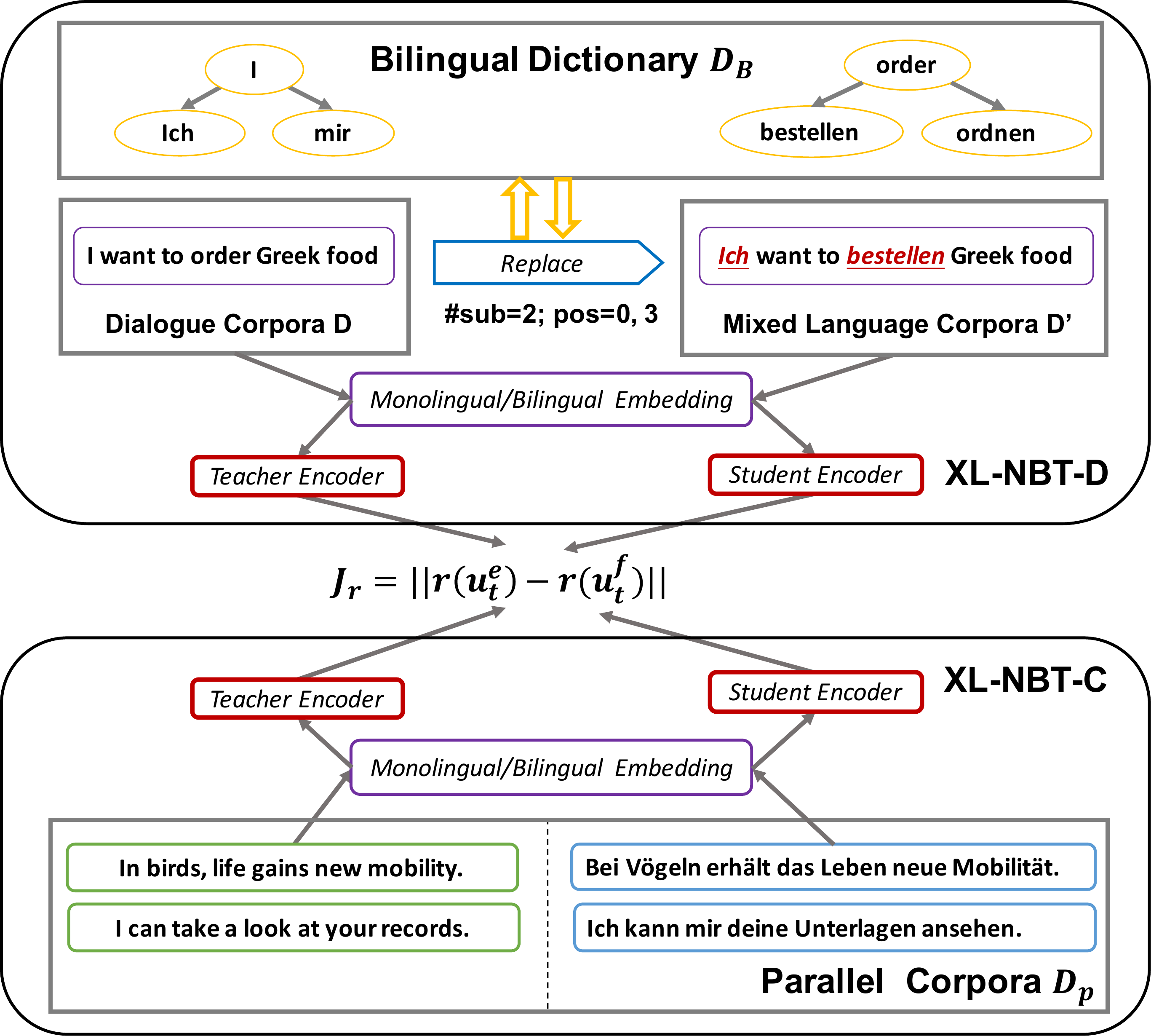}
\end{center}
\vspace*{-2ex}   
\caption{XL-NBT-C and XL-NBT-D for two scenarios}
\label{fig:Sent-Level}
\vspace*{-2ex}   
\end{figure}

We are given a well-trained NBT for a source language $e$, and we want to learn an NBT for a target language $f$ without any annotated training data. Therefore, we cannot learn the target-side NBT from standard supervised learning. Instead, we use a teacher-student framework to distill the knowledge from the source-side NBT (teacher  network) into the target-side NBT (student network) (see~\autoref{fig:teacher-student}). Let $x_e \defeq (c^e_s, c^e_v,u^e_t, a^e_t)$ be the input to the teacher network and let $x_f \defeq (c^f_s, c^f_v,u^f_t, a^f_t)$ be the associated input to the student network. The standard teacher-student framework trains the student network by minimizing
\begin{align}
\label{eq:objective}
\small
\begin{split}
    \hat{J_1} = 
    \sum_{x_e, x_f} 
    ||y(c^e_s, c^e_v, u^e_t, a^e_t) - y(c^f_s, c^f_v, u^f_t, a^f_t)||^2
\end{split}
\end{align}
\vspace{-0ex}%
where $y(c^e_s, c^e_v, u^e_t, a^e_t)$ and $y(c^f_s, c^f_v, u^f_t, a^f_t)$ denote the scores by the teacher and the student networks, respectively, and the slot-value pairs satisfy $M(c^f_v)=M(c^e_v)$ and $M(c^f_s)=M(c^e_s)$. However, the target-side inputs $(c^f_s, c^f_v, u^f_t, a^f_t)$ parallel to $(c^e_s, c^e_v, u^e_t, a^e_t)$ are usually not available in cross-lingual DST, and, even worse, the target-side utterance $u_t^e$ is not available. We may have to generate synthetic input data for the student network or leverage external data sources. It is relatively easy to use the mapping $M(\cdot)$ to generate $(c_s^f,c_v^f,a_t^f)$) (i.e., the inputs of the target-side context gate) from the $(c_s^e,c_v^e,a_t^e)$. But it is more challenging to obtain the parallel utterance data $u_t^f$ from $u_t^e)$. Therefore, we have to leverage external bilingual data sources to alleviate the problem. However, the external bilingual data are usually not in the same domain as the utterance, and hence they are not aligned with the slot-value pair and system acts (i.e., $(c_s^e,c_v^e,a_t^e)$ or $(c_s^f,c_v^f,a_t^f)$). For this reason, we cannot perform the knowledge transfer by minimizing the cost \eqref{eq:objective}. Instead, we need to develop a new cost function where the utterance is not required to be aligned with the slot-value pair and the system acts. To this end, let $g^e = g_e(c^e_s, c^e_v, a^e_t)$ and $g^f = g_f(c^f_s, c^f_v, a^f_t)$. And we substitute~\eqref{eq:decompose} into \eqref{eq:objective} and get:
\begin{align*}
\small
\begin{split}
   \hat{J_1} 
            &\le 
                    ||W_y||^2\sum_{c^f_v,c^e_v} ||r_e \odot g^e - r_f \odot g^f||^2
                    \\
            &=  
                    ||W_y||^2\sum_{c^f_v,c^e_v} ||g^e \odot (r_e - r_f) + r_f \odot (g^e - g^f)||^2\\ 
            &\le 
                    ||W_y||^2\sum_{c^f_v,c^e_v} ||g^e||^2 ||r_e - r_f||^2  + ||r_f||^2 ||g^e - g^f||^2
\end{split}
\end{align*}
where $r_e = r_e(u_t^e)$ and $r_f = r_f(u_t^f)$.
As we mentioned earlier, the weight $W_y$ in the slot-value decoder is shared between the student and the teacher networks and will not be updated. The teacher-student optimization only adjusts the weights related to the language-specific parts in Figure \ref{fig:NBT} (i.e., utterance encoding and context gating). Therefore, the shared weight $||W_y||$ is seen as a constant.
Furthermore, $\sum_{c^f_v,c^e_v} ||g^e||^2$ can be seen as a constant since the teacher gate is fixed. Since we use batch normalization layer to normalize the encoder output (described in~\autoref{fig:NBT}), $||r_f(u_t^f)||^2$ can also be treated as a constant $C_2$. Therefore, we formally write the upper bound of $\hat{J_1}$ as our surrogate cost function $J$:
\begin{align}
\small
\begin{split}
    J =& 
            C_1 ||r_e(u_t^e) - r_f(u_t^f)||^2
            + 
            C_2 \sum_{c^f_v,c^e_v} ||g^e - g^f||^2
\end{split}
\label{Equ:SurrogateCost}
\end{align}
The surrogate cost has successfully decoupled utterance encoder with context gate, and we use $J_r$ and $J_g$ to measure the encoder matching cost and the gate matching cost, respectively.
\begin{align}
\small
    \begin{split}
        J_r &= ||r_e(u_t^e) - r_f(u_t^f)||^2\\
        J_g &= \sum_{c^f_v,c^e_v} ||g^e - g^f||^2
    \end{split}
\end{align}
The encoder cost $J_r$ is optimized to distill the knowledge from the teacher encoder to student encoder while gate cost $J_g$ is optimized to distill the knowledge from teacher gate to student gate. This objective function successfully \emph{decouples} the optimization of encoder and gate, thus we are able to optimize $J_r$ and $J_g$ separately from different data sources. Recall that we can easily simulate the target-side system acts, slot-value pairs $(c_s^f, c_v^f, a^f)$  by using the ontology mapping $M$. Therefore, optimizing $J_g$ is relatively easy. Formally, we write the gate matching cost as follows:
\begin{align}
\small
\begin{split}
J_g = \sum_{\substack{a^e_t, c^e_s, c^e_v \\  a^f_t, c^f_s, c^f_v}} ||g_e(c^e_s,c^e_v,a^e_t) - g_f(c^f_s,c^f_v,a^f_t)||^2
\end{split}
\label{Equ:J_g}
\end{align}
However, exact optimization of $J_r$ is difficult and we have to approximate it using external parallel data. We consider two kinds of external resources (bilingual corpus and bilingual dictionary) in the sections \ref{Sec:XL_NBT_C}-\ref{Sec:XL_NBT_D} (see ~\autoref{fig:Sent-Level} for the main idea).

\subsection{Bilingual Corpus (XL-NBT-C)}
\label{Sec:XL_NBT_C}
In our first scenario, we assume there exists a parallel corpus $D_p$ consisting of sentence pairs from the source language and the target language. In this case, the cost function \eqref{Equ:SurrogateCost} is approximated by
\begin{align}
\small
\begin{split}
    J = \expect{(m_e, m_f) \in D_p} ||r_e(m_e) - r_f(m_f)||^2 + \alpha J_g
\end{split}
\label{Equ:J_XL_NBT_C}
\end{align}
where $\alpha$ is the balancing factor and $J_g$ is defined in \eqref{Equ:SurrogateCost}. The cost function \eqref{Equ:J_XL_NBT_C} is minimized by stochastic gradient descent. At test time, we switch the encoder to receive target language inputs.

\subsection{Bilingual Dictionary (XL-NBT-D)}
\label{Sec:XL_NBT_D}
In the second scenario, we assume there exists no parallel corpus but a bilingual dictionary $D_B$ that defines the correspondence between source words and target words (a one-to-many mapping $\{w: M_D(w)\}$). Likewise, it is infeasible to optimize the exact encoder cost $J_r$ due to the lack of target-side utterances. We propose a word replacement strategy (to be described later) to generate synthetic parallel sentence $\hat{u}_t^f$ of ``mixed" language. Then, we use the generated target parallel sentences to approximate the cost \eqref{Equ:SurrogateCost} by
\begin{align}
\small
\begin{split}
    J_r = \expect{u_t \in D} ||r_e(u^e_t) - r_f(\hat{u}_t^f)||^2 + \alpha J_g
\end{split}
\label{Equ:J_XL_NBT_D}
\end{align}
where $\alpha$ is the balancing factor. For word replacement, we first decide the number of words $N_w$ to be replaced, then we draw $N_w$ positions randomly from the source utterance and substitute the corresponding word $w_i$ with their target word synonym from $M_D(w)$ based on the context as follows:
\begin{align}
\small
\begin{split}j
    p(N_w=i) &= \frac{ \exp(-i / \tau)}{\sum_{i' < N} \exp(- i' / \tau)}\\
    p(\hat{w}) &= \frac{\hat{w} \cdot h_{\hat{w}}}{\sum_{w'  \in M(w_i)} w' \cdot h_{\hat{w}}}\\
\end{split}
\end{align}
where $h_{\hat{w}} = \sum_{k=-2: k \neq 0}^2 w_{i+k}$ represents the context vector and $N$ denotes the utterance length. The context similarity of context and the target-side synonym can better help us in choosing the most appropriate candidate from the list. In our following experiments, we adjust the temperature of $\tau$ to control the aggressiveness of replacement. 

\section{Experiments}
\label{Sec:Experiments}
\subsection{Dataset}
The Wizard of Oz (WOZ)~\cite{DBLP:conf/eacl/Rojas-BarahonaG17} dataset is used for training and evaluation, which consists of user conversations with task-oriented dialog systems designed to help users find suitable restaurants around Cambridge, UK. The corpus contains three informable (i.e. goal-tracking) slots: FOOD, AREA, and PRICE. The users can specify values for these slots in order to find which best meet their criteria. Once the system suggests a restaurant, the users can ask about the values of up to eight requestable slots (PHONE NUMBER, ADDRESS, etc.).  Multilingual WOZ 2.0~\cite{DBLP:conf/acl/MrksicSWTY17} has expanded this dataset to include more dialogs and more languages. The train, valid and test datasets for three different languages (English, German, Italian) are available online\footnote{\small\url{https://github.com/nmrksic/neural-belief-tracker/tree/master/data}}. We use the English as source language where 600 dialogs are used for training, 200 for validation and 400 for testing. We use the German and Italian as the target language to transfer our knowledge from English DST system. In the experiments, we do not have access to any training or validation dataset for German and Italian, and we only have access to their testing dataset which is composed of 400 dialogs. 

For external resource, we use the IWSLT2014 Ted Talk parallel corpus~\cite{mauro2012wit3} from the official website\footnote{\small\url{https://wit3.fbk.eu/mt.php?release=2014-01}} for bilingual corpus scenario. In the IWSLT2014 parallel corpus, we only keep the sentences between 4 and 40 words and decrease the sentence pairs to around 150K. We use Panlex~\cite{kamholz2014panlex} as our data source and crawl translations for all the words appearing in the dialog datasets to build our bilingual dictionary. We specifically investigate two kinds of pre-trained embedding, and we use Glove~\cite{pennington2014glove} as the monolingual embedding and MUSE~\cite{DBLP:journals/corr/abs-1710-04087} as the bilingual embedding to see their impacts on the DST performance.

We split the raw DST corpus into turn-level examples. During training, we use the ground truth previous state $V_{t-1}$ as inputs. At test time, we use the model searched states as the previous state to continue tracking intention until the end of the dialog. When the dialog terminates, we use two evaluation metrics introduced in~\citet{henderson2014robust} to evaluate the DST performance: (1) \textit{Goals}: the proportion of dialog turns where all the user’s search goal constraints were correctly identified. (2) \textit{Requests}: similarly, the proportion of dialog turns where user’s requests for information were identified correctly. Our implementation is based on the NBT\footnote{\small \url{https://github.com/nmrksic/neural-belief-tracker}}, the details of our system setting are described in the appendix. 

\begin{table*}[htb]
\small
\centering
\begin{tabular}{c|l}
\toprule
\textbf{Error Type} & \multicolumn{1}{c}{\textbf{Examples}}\\ \midrule
\begin{tabular}[c]{@{}c@{}}Modify \\ Failure\end{tabular}                    & \begin{tabular}[c]{@{}l@{}}Machine: I have two options that fit that description, golden wok Chinese restaurant and the Nirala which \\ serves Indian food,  do you have a preference?\\ User: How about Nirala, what’s the address and phone of that?\\ Previous State: food=Chinese; Prediction: food=none; Groundtruth: food=Indian\end{tabular} \\ \midrule
\begin{tabular}[c]{@{}c@{}}Maintain \\ Failure\end{tabular} & \begin{tabular}[c]{@{}l@{}}Machine: there are \$num places with a moderate price range. can you please tell me what kind of food\\ you would like?\\ User: well I want to eat in the north, what’s up that way?\\ Previous State: food=expensive; Prediction: food=none; Groundtruth: food=expensive\end{tabular} \\ \midrule
\begin{tabular}[c]{@{}c@{}}History \\ Failure\end{tabular}                   & \begin{tabular}[c]{@{}l@{}}Machine: Anatolia is located at \$num bridge street city center.\\ User: thank you goodbye!\\ Previous State: food=Chinese; Prediction: food=Chinese;,Groundtruth: food=Turkish\end{tabular} \\ \bottomrule
\end{tabular}
\label{tab:error-type}
\caption{Here we show the frequent error types, the examples are translated to English for better understanding.}
\end{table*}

\begin{table*}[tbh]
\centering
\small
\scalebox{1.0}{
\begin{tabular}{l|lllllll}
\toprule
    \multicolumn{2}{c}{\textbf{Language}}  & \multicolumn{2}{c}{\textbf{German (student)}} & \multicolumn{2}{c}{\textbf{Italian (student)}} & \multicolumn{2}{c}{\textbf{English (teacher)}} \\
\midrule
    \multicolumn{2}{c}{Models} & Goal    & Request & Goal   & Request        & Goal               & Request           \\
\midrule
\multirow{3}{*}{\textit{w/ Supervised Dialog}}& NBT~\cite{DBLP:conf/acl/MrksicSWTY17}      & -   & -  & -    & -  & \textbf{0.84} & \textbf{0.91}              \\
& Decoupled NBT (mono)        & 0.79              & 0.83              & 0.86               & 0.91              & 0.82               & 0.89              \\ 
& Decoupled NBT (bilingual)   & \textbf{0.80}     & \textbf{0.84}     & \textbf{0.88}      & \textbf{0.91}     & \textbf{0.84}      & 0.90     \\
\midrule
\multirow{3}{*}{\textit{w/o Bilingual Data}} & w/o Transfer (mono)      & 0.15              & 0.10              & 0.15               & 0.11              & -                  & -                 \\
                                         & w/o Transfer (bilingual) & 0.13              & 0.13              & 0.11               & 0.12              & -                  & -                 \\ 
                                         & Ontology Matching        & 0.24              & 0.21              & 0.23               & 0.21              & -                  & -                 \\
\midrule
    \multirow{3}{*}{\textit{w/ Bilingual Corpus}}     & Translate~\cite{2017opennmt}                & 0.41              & 0.42              & 0.48               & 0.51 &                                      -                  & -                 \\
                                         & XL-NBT-C (mono)          & 0.48              & 0.54              & 0.65               & 0.60              & -                  & -                 \\
                                         & XL-NBT-C (bilingual)     & \textbf{0.55}     & \textbf{0.59}     & \textbf{0.72}      & \textbf{0.69}     & -                  & -                 \\
\midrule
    \multirow{3}{*}{\textit{w/ Bilingual Dictionary}} & Word-by-Word             & 0.22              & 0.25              & 0.25               & 0.27& 
                                         -                  & -                 \\ 
                                         & XL-NBT-D (mono)          & 0.14              & 0.15              & 0.23               & 0.22              & -                  & -                 \\
                                         & XL-NBT-D (bilingual)     & \textbf{0.51}     & \textbf{0.56}     & \textbf{0.73}      & \textbf{0.63}     & -                  & -                 \\ 
\bottomrule
\end{tabular}}
\vspace*{-1ex}   
\caption{Experimental results for cross-lingual NBT and other baseline algorithms. All results are averaged over 5 runs. Here we use ``mono" to refer to the experiments with pre-trained monolingual embedding, ``bilingual" to refer to the experiments with pre-trained bilingual embedding. }
\vspace*{-2ex}   
\label{tab:results}
\end{table*}

\nop{\begin{figure}[th]
\begin{center}
\includegraphics[width=1.0\linewidth]{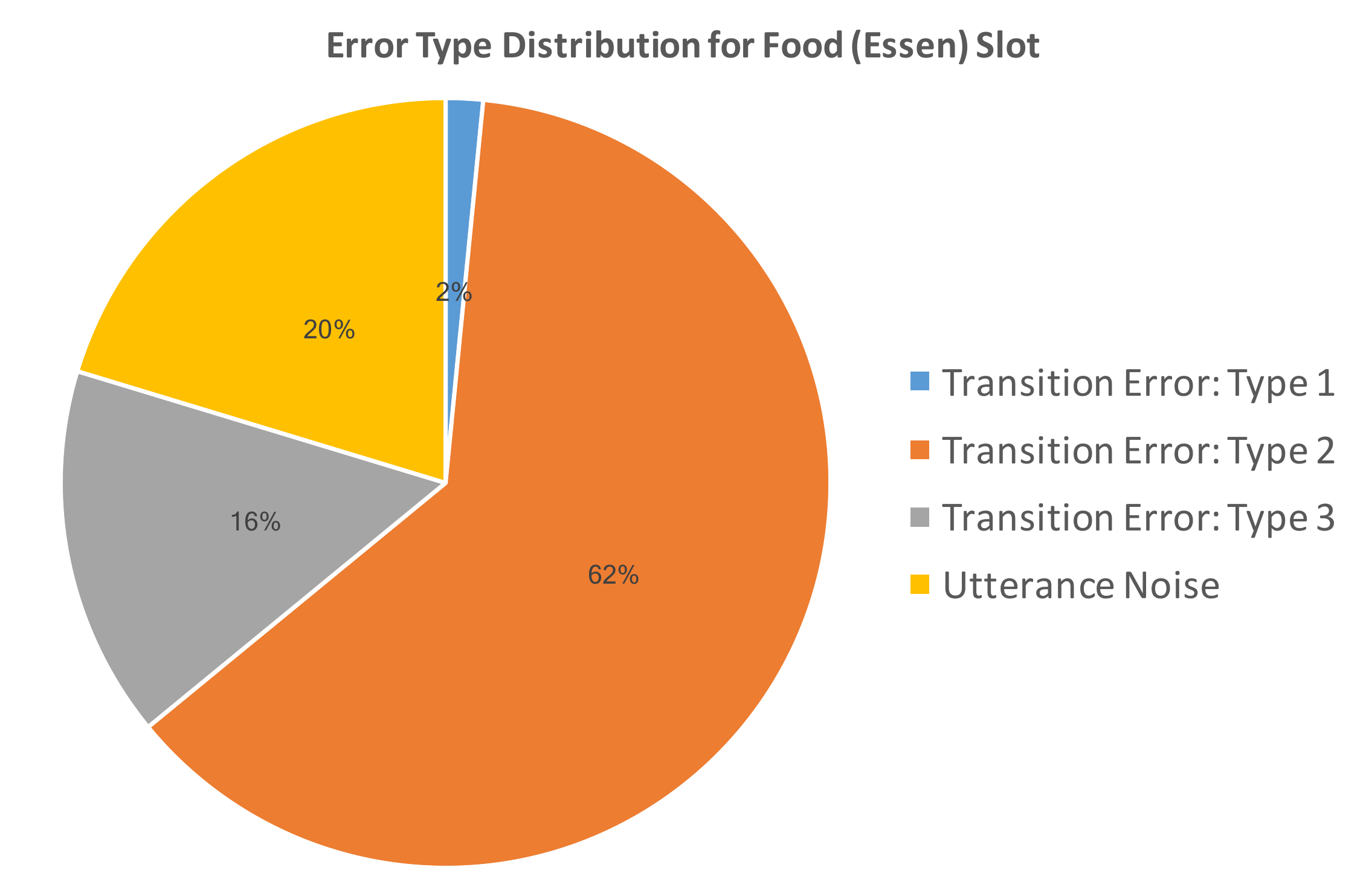}
\end{center}
\vspace*{-2ex}   
   \caption{The distribution of different kinds of errors in sentence-level transfer}
\label{fig:pie-chart}
\vspace*{-2ex}   
\end{figure}
}
\subsection{Results}
Here we highlight the baselines we use to compare with our cross-lingual algorithm as follows:\\
\noindent \textbf{(1) Supervised}: this baseline algorithm assumes the existence of annotated dialog belief tracking datasets, and it determines the upper bound of the DST model.\\
\noindent \textbf{(2) w/o Transfer}: this algorithm trains an English NBT, and then directly feeds target language into the embedding level as inputs during test time to evaluate the performance.\\
\noindent \textbf{(3) Ontology-match}: this algorithm directly uses exact string matching against the utterance to discover the perceived slot-value pairs, it directly assigns a high score to the appearing candidates.\\
\noindent \textbf{(4) Translation-based}: this system pre-trains a translator on the external bilingual corpus and then translates the English dialog and ontology into target language as ``annotated" data, which is used to train the NBT in the target language domain (more details about the implementation, performance and examples are listed in the appendix).\\
\noindent \textbf{(5) Word-By-Word (WBW)}: this system transforms the English dialog corpus into target language word by word using the bilingual dictionary, which is used to train the NBT in target side.\\
We demonstrate the results for our proposed algorithms and other competing algorithms in~\autoref{tab:results}, from which we can easily conclude that that (i) our Decoupled NBT does not affect the performance, and (ii) our cross-lingual NBT framework is able to achieve significantly better accuracy for both languages in both parallel-resource scenarios.

\paragraph{Compare with Translator/WBW.}
With bilingual corpus, XL-NBT-C with pre-trained bilingual embedding can significantly outperform our Translator baseline~\cite{2017opennmt}. This is intuitive because the translation model requires both source-side encoding and target-side word-by-word decoding, while our XL-NBT only needs a bilingual source-encoding to align two vector space, which averts the compounded decoding errors. With the bilingual dictionary, the word-by-word translator is very weak and leading to many broken target sentences, which poses challenges for DST training. In comparison, our XL-NBT-D can control the replacement by adjusting its temperature to maintain the stability of utterance representation. Furthermore, for both cases, our teacher-student framework can make use of the knowledge learned in source-side NBT to assist its decision making, while translator-based methods learn from scratch.

\paragraph{Bilingual Corpus vs. Bilingual Dictionary.}
From the table, we can easily observe that bilingual corpus is obviously a more informative parallel resource to perform cross-lingual transfer learning. The accuracy of XL-NBT-D is lower than XL-NBT-C. We conjecture that our replacement strategy to generate ``mixed" language utterance can sometimes break the semantic coherence and cause additional noises during the transfer process, which remarkably degrades the DST performance.

\paragraph{Monolingual vs. Bilingual embedding.}
From the table, we can observe that the bilingual embedding and monolingual embedding does not make much difference in supervised training. However, the gap in the bilingual corpus case is quite obvious. Monolingual embedding even causes the transfer to fail in a bilingual dictionary case. We conjecture that the bilingual word embedding already contain many alignment information between two languages, which largely eases the training of encoder matching objective.

\paragraph{German vs. Italian}
As can be seen, the transfer learning results for Italian are remarkably higher than German, especially for the ``Goal'' accuracy. We conjecture that it is due to German declension, which can produce many word forms. The very diverse word forms present great challenges for DST to understand its intention behind. Especially for the bilingual dictionary, German tends to have much longer replacement candidate lists than Italian, which introduces more noises to the replacement procedure. 

\paragraph{Error Analysis}
Here we showcase the most frequent error types in~\autoref{tab:error-type}. From our observation, these three types of errors distribute evenly in the test dialogs. The error mainly comes from the unaligned utterance space, which leads to failure in understanding the intention of human utterance in the target language. This can lead the system to fail in modifying the dialog state or maintaining the previous dialog states. 

\subsection{Discussion}
Here we want to further highlight the comparison between our transfer learning algorithm with the MT-based approach. Though our approach outperforms the standard Translator trained on IWSLT-2014, it does not necessarily claim that our transfer algorithm outperforms any translation methods on any parallel corpus. In our further ablation studies, we found that using Google Translator~\footnote{\url{https://translate.google.com/}} can actually achieve a better score than our transfer algorithm, which is understandable considering the complexity of Google Translator and the much larger parallel corpus it leverages. By leveraging more close-to-domain corpus and comprehensive entity recognition/replacement strategy, the translator model is able to achieve a higher score. Apparently, we need to trade off the efficiency for the accuracy. For DST problem, it is an overkill to introduce a more complex translation algorithm, what we pursue is a simple yet efficient algorithm to achieve promising scores. It is also worth mentioning that our XL-NBT algorithm only takes several hours to achieve the reported score, while the translator model takes much more time and memory to train depending on the complexity. Thus, the simplicity and efficiency makes our model a better fit for rare-language and limited-budget scenarios.

\subsection{Ablation Test}
Here we investigate the effect of hyper-parameter $\alpha, \tau$ on the evaluation results. The $\alpha$ is used to balance the optimization of encoder constraint and gate constraint, where larger $\alpha$ means more optimization on gate constraint. The temperature $\tau$ is used to control the aggressiveness of the replacement XL-NBT-D, where smaller $\tau$ means more source words are replaced by target synonyms.
\begin{table}[htb]
\small
    \centering
    \begin{tabular}{llllll}
    \toprule
        \multicolumn{3}{c}{\textbf{$\boldsymbol{\alpha}$ ablation ($\boldsymbol{\tau}$ fixed to 0.1)}} & \multicolumn{3}{c}{\textbf{$\boldsymbol{\tau}$ ablation ($\boldsymbol{\alpha}$ fixed to 1)}} \\
    \midrule
        value  & Goal & Request & value & Goal & Request \\ 
    \midrule
        $\alpha$=0 & 0.13 & 0.00 & $\tau$=0 & 0.14 & 0.08 \\
        $\alpha$=0.1 & 0.46 & 0.54 & $\tau$=0.03 & 0.43 & 0.50 \\
        $\alpha$=1 & \textbf{0.51} & \textbf{0.56} & $\tau$=0.1 & \textbf{0.51} & \textbf{0.56} \\
        $\alpha$=5 & 0.46 & 0.54 & $\tau$=0.3 & 0.47 & 0.51\\
        $\alpha$=10 & 0.46 & 0.52 & $\tau$=1 & 0.44 & 0.52 \\
        $\alpha$=100 & 0.44 & 0.50 & $\tau$=10 & 0.33 & 0.32 \\
    \bottomrule
    \end{tabular}
    \vspace*{-1ex}   
    \caption{Ablation test for hyper-parameter $\alpha$ and $\tau$ on English-to-German XL-NBT-D.}
    \vspace*{-2ex}   
    \label{tab:ablation}
\end{table}
From the table~\autoref{tab:ablation}, we can observe that the experimental results are not very sensitive to $\alpha$, a dramatic change of $\alpha$ will not harm the final results too much, we simply choose $\alpha=1$ as the hyper-parameter. In contrast, the system is more sensitive to temperature. Too conservative replacement will lead to weak transfer, while too aggressive replacement will destroy the utterance representation. Therefore, we choose the a moderate temperature of $\tau=0.1$ throughout our experiments. We also draw the learning curve (Precision vs. Iteration) in the Appendix for both XL-NBT-C and XL-NBT-D. The learning curves show that our algorithm is stable and converges quickly, and the reported results are highly reproducible. 

\section{Conclusion}
In our paper, we propose a novel teacher-student framework to perform cross-lingual transfer learning for DST. The key idea of our model is to decouple the current DST neural network into two separate modules and transfer them separately. We believe our method can be further extended into a general purpose multi-lingual transfer framework to resolve other NLP matching or classification problems. 
\section{Acknowledgement}
We are gratefully supported by a Tencent AI Lab Rhino-Bird Gift Fund. We are also very thankful for the public belief tracking code and multi-lingual state-tracking datasets released by Nikola Mrksic from the University of Cambridge. 

\clearpage
\bibliography{emnlp2018}
\bibliographystyle{acl_natbib_nourl}

\clearpage
\appendix

\section{Supplemental Material}
\subsection{Translator Baseline}
In this paper, we use translator model as baseline to compare with our cross-lingual transfer algorithm. Here we adopt the open-source toolkit for neural machine translation (NMT) in openNMT\footnote{https://github.com/OpenNMT/OpenNMT-py} and train this model on the IWSLT2014 english-to-german and english-to-italian translation dataset. Our training data consists of 178K sentence pairs and validation data consists of 2K sentence pairs. We train the NMT model for roughly five epochs until it converges on the validation dataset. The source and target word vector size are both set to 500, the attention-based RNN hidden size for encoder  and decoder are also both 500. We adopt the input feeding strategy to further boost the performance and apply beam size of 8 to search for the most probable target candidate. We leave the UNKs out during beam search to reduce noises.
Here we first show the results of our trained model against some state-of-art algorithms to verify that our trained translation model makes sense:
\begin{table}[htb]
    \centering
    \begin{tabular}{l|c}
    \hline
         Model & BLEU  \\
    \hline
         Actor-Critic~\cite{bahdanau2016actor} & 28.53 \\
         BSO~\cite{wiseman2016sequence} & 26.63 \\
         Variational Attention~\cite{deng2018latent} & 33.09\\
         Our OpenNMT~\cite{2017opennmt} & 30.2 \\
    \hline
    \end{tabular}
    \caption{The test set results on IWSLT for different algorithms}
    \label{tab:translator}
\end{table}
We can see from~\autoref{tab:translator} that our pre-trained model is able to achieve reasonable results on the common translation dataset. 
Here we demonstrate some examples for translated dialogs:
\begin{itemize}
    \item Human(\textbf{correct}): gibt es irgendwelche eritrean restaurants in die stadt? (is there any eritrean restaurants in the town?)
    \item Machine: no, there are no eritrean restaurants in town. would you like a different restaurant?
    \item Human(\textbf{missing}): Wie steht es mit essen? (Is there any \st{Chinese} food?)
    \item Machine: there is a wide variety of chinese restaurants, do you have an area preference or a price preference to narrow it down?
    \item Human(\textbf{correct}): ich würde den osten der stadt mögen. (I want to try something in the east)
    \item Machine: yu garden is a chinese restaurant in the east area.
    \item Human(\textbf{correct}): könnte ich die adresse bekommen, und postcode garden? (Can I obtain the address and postcode of the restaurant?)
    \item Machine: phone is xxx-xxxx, address and postcode are xxx new market road
\end{itemize}
Another example is shown below:
\begin{itemize}
    \item Human(\textbf{missing}): ich möchte ein restaurant in jedem gebiet finden , und serves essen? (I want to find a restaurant anywhere to serve \st{Scottish} food?)
    \item Machine: there are no restaurants serving scottish food . would you like to try something else?
    \item Human(\textbf{correct}): ja, wie steht es mit modernen europäischen essen? (how about moderate european food?)
    \item Machine: eraina is a great european eatery. what other information can i provide to you?
    \item Human(\textbf{correct}): was ist das telefon? (What is its telephone?)
    \item Machine: their phone number is $num $num may i help you with anything else?
    \item Human(\textbf{correct}): nein, vielen dank. (no, thanks.)
\end{itemize}
The above generated German dialogs are shown with their English translation to help readers to understand. We have observed that the most frequent two error types are the entity missing and broken grammar. Since IWSLT2014 lies in another domain, its vocabulary is mis-aligned with the spoken dialog vocabulary. Therefore, the NMT decoder fails to generate many ontology words. These errors in the translated dialog corpus can remarkably degrade the DST performance.
\subsection{NBT system setting}
In our implementation, we set the word embedding size to be $H=300$ for both monolingual and bilingual embedding. We directly use the ontology's corresponding lexicon embedding as its own embedding. Therefore, the ontology embedding also shares the same dimension. We restrict the length of utterance to be a maximum of 40 and filter the utterances beyond that length during training. Our CNN has three independent filter size (1, 2, 3) to extract unigram, bigram and trigram information out of the utterance. The intermediate feature dimension of CNN is also set to 300, finally we add the three filters to construct the utterance representation. We also apply dropout strategy after the three gates.
\subsection{Learning Curve}
\begin{figure}[hbt]
    \begin{center}
    \includegraphics[width=1.0\linewidth]{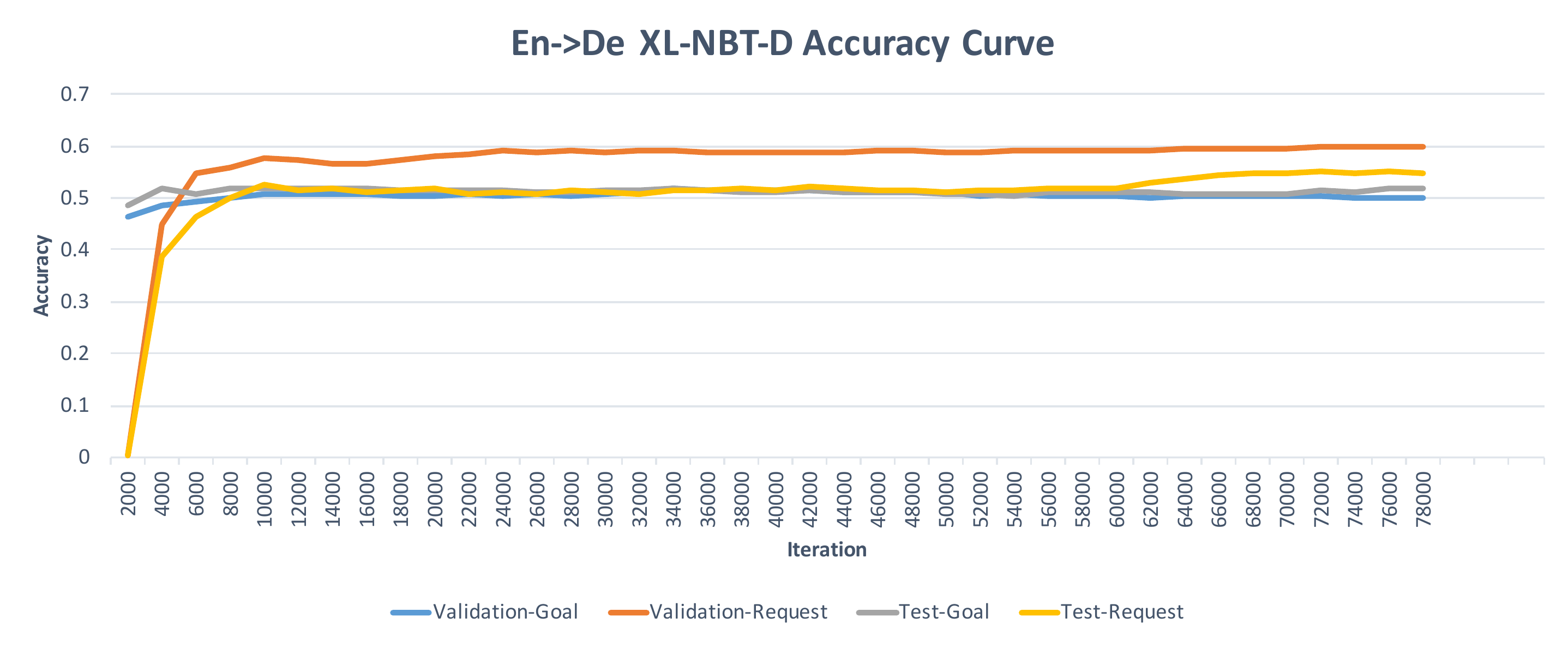}
    \end{center}
    \vspace*{-2ex}
    \caption{The learning curve for Transfer Learning (XL-NBT-D).}
    \label{fig:learning_curve_d}
    \vspace*{-2ex}
\end{figure}

\begin{figure}[hbt]
    \begin{center}
    \includegraphics[width=1.0\linewidth]{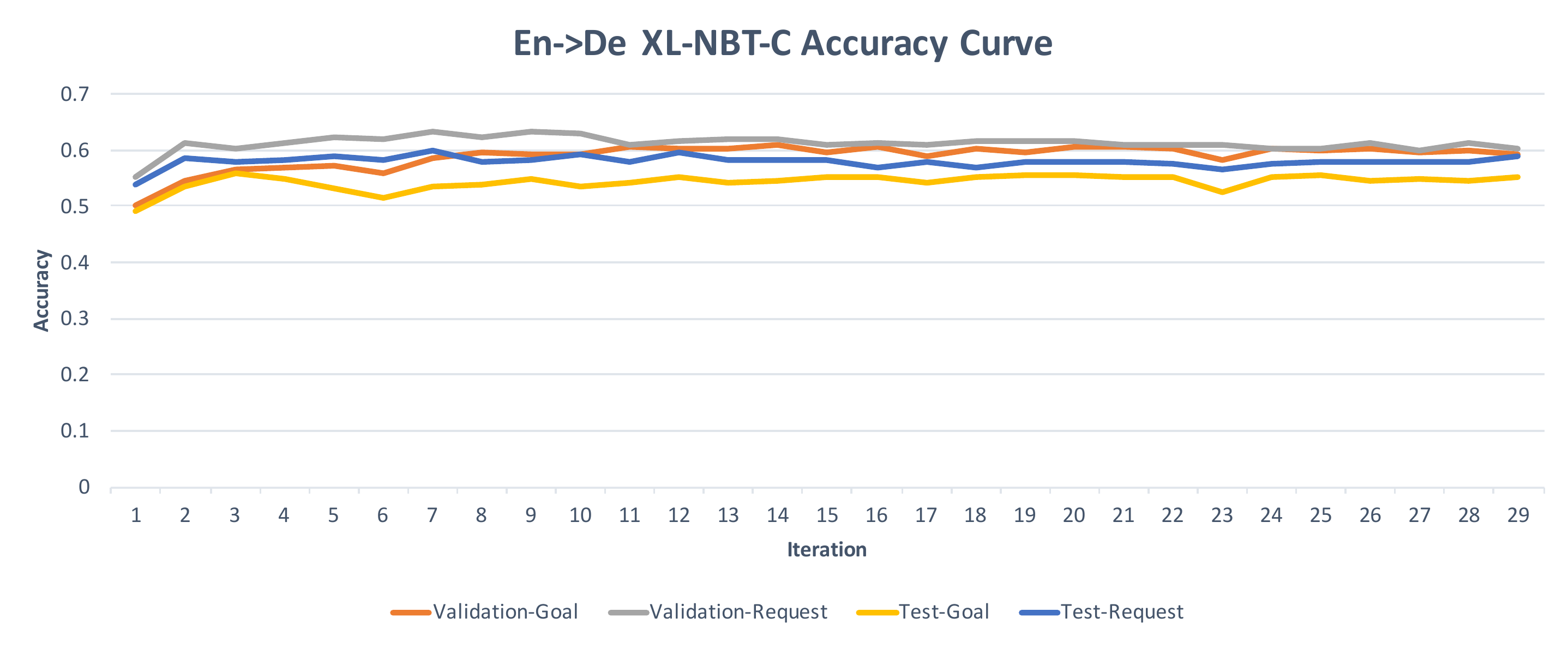}
    \end{center}
    \vspace*{-2ex}
    \caption{The learning curve for Transfer Learning (XL-NBT-C).}
    \label{fig:learning_curve_c}
    \vspace*{-2ex}
\end{figure}

Here we demonstrate the learning curve for XL-NBT-D in~\autoref{fig:learning_curve_d} and XL-NBT-C in~\autoref{fig:learning_curve_c}.
The rise of our transfer learning is very steady, we average multiple runs as our final reported score in the paper.
\end{document}